\documentclass[10pt, a4paper]{article}

\usepackage{lrec-coling2024} 

\title{Making a prototype of Seoul historical sites chatbot using Langchain}

\name{Jae Young Suh, Minsoo Kwak, Soo Yong Kim, Hyoungseo Cho} 

\address{Hanyang, Konkuk, Seoul National, Myongji University \\
         tjwodud04@gmail.com, 
         estherpd@naver.com, 
         ksyint1111@snu.ac.kr, 
         hyoungseocho@gmail.com\\
         }

\abstract{
In this paper, we are going to share a draft of the development of a conversational agent created to disseminate information about historical sites located in the Seoul.The primary objective of the agent is to increase awareness among visitors who are not familiar with Seoul, about the presence and precise locations of valuable cultural heritage sites. It aims to promote a basic understanding of Korea’s rich and diverse cultural history. The agent is thoughtfully designed for accessibility in English and utilizes data generously provided by the Seoul Metropolitan Government. Despite the limited data volume, it consistently delivers reliable and accurate responses, seamlessly aligning with the available information. We have meticulously detailed the methodologies employed in creating this agent and provided a comprehensive overview of its underlying structure within the paper. Additionally, we delve into potential improvements to enhance this initial version of the system, with a primary emphasis on expanding the available data through our prompting. In conclusion, we provide an in-depth discussion of our expectations regarding the future impact of this agent in promoting and facilitating the sharing of historical sites.
\\ \newline \Keywords{Conversational agent, Historical sites} 
}

\begin{document}


\maketitleabstract

\section{Introduction}
We've become captivated by how culture and technology blend together, leading us to create a chatbot focused on exploring Seoul's historic sites in Korea. Seoul is filled with cultural gems waiting to be explored, offering deep insights into Korean heritage. The city's top-notch public transport and convenience showcase the many aspects of Korea.

In this paper, we'll talk about how we built this chatbot. We used tools like Streamlit, Langchain, and the OpenAI API. Streamlit helped us make an interactive web app that's easy to use. Langchain was key in making the chatbot understand and respond more naturally.

Next, we'll share how we developed the chatbot, blending technology with information to create a reliable guide for cultural discovery. We'll explain how we gathered and organized information from public sources, made sure the chatbot's conversations flowed logically, and polished the user interface to keep users engaged without interruptions.

Then, we'll discuss the chatbot's benefits, focusing on how it can help people learn about Korean culture, especially those new to it. We'll also talk about the challenges we faced, like limitations due to the amount of available data, and our plans to add more information.

We'll wrap up by reflecting on what this project means, like how technology can help us connect with and appreciate different cultures. We'll also write about future improvements, hoping to create a more comprehensive and immersive experience that covers a wide range of cultural aspects.

Our goal with this paper is to shed light on the process of creating a chatbot enriched with culture and to spark discussions about using technology for wider and more inclusive cultural explorations.

\section{Methods}
\subsection{Langchain}

At the heart of our heritage-focused conversational agent lies Langchain, the primary infrastructure for driving our natural language processing (NLP) functionalities. Langchain is a robust and efficient framework designed specifically for building conversational AI models. This enables the creation of the agent that are not only responsive but also highly skilled in comprehending and generating natural and fluent interactions.

In our endeavor, prowess of Langchain was unmistakably manifested as it gave life to the agent adept in navigating the cultural tapestry of Seoul's heritage sites. This framework conferred the agent with an acute sense of understanding, transforming user queries into coherent, context-rich dialogues. Beyond mere response generation, Langchain’s architecture delves deep into the intricacies of conversation, mastering user intent recognition, context management, and eloquent response formulation.

\subsection{Streamlit}

Streamlit is a prominent tool in web application development, particularly when integrating data science and machine learning. As an open-source library built on Python, Streamlit is distinguished by its straightforwardness, enabling swift and efficient creation and launch of web applications.

Our application utilized Streamlit to develop a visually appealing and interactive interface for the agent. Streamlit efficiently integrated the natural language processing capabilities provided by Langchain. As a result, the platform offers real-time conversations, easy-to-use input fields, and an engaging display. Essentially, Streamlit enhanced the interface of the system, making it user-friendly and improving the overall user experience.

The combination of Langchain’s advanced conversational AI capabilities with Streamlit’s web development and user-friendly.

\subsection{OpenAI API}

The OpenAI API, an essential tool in today's AI development, was crucial in enhancing capabilities of the agent. With the help of advanced language models like GPT-4 from this API, created answers are much more natural and fluent. As a result, the agent doesn't just give basic information, but it offers detailed and thoughtful responses.

We made significant use of the API, taking full advantage of its advanced language understanding. If users wrote their question to the chatbot, the API carefully analyzed and used this information to create a tailored answer. Additionally, the ability of the API let the agent to understand and communicate in multiple languages which it can cater various input available, making them feel welcomed and understood.

By integrating the OpenAI API into our project, we've developed a digital cultural chatbot that can understand different languages and provide tailored experiences for users.

\section{Structure}

In this section, we will explain how we built the conversational agent. Its main purpose is to provide accurate information about historical sites in Seoul, especially their names and brief location details. This information is sourced from a specially prepared data file that lists Seoul's heritage in English(\citetlanguageresource{Seoul}). The functioning of the agent can be visualized in the image that follows.

\begin{figure}[!ht]
\begin{center}
\includegraphics[scale=0.265]{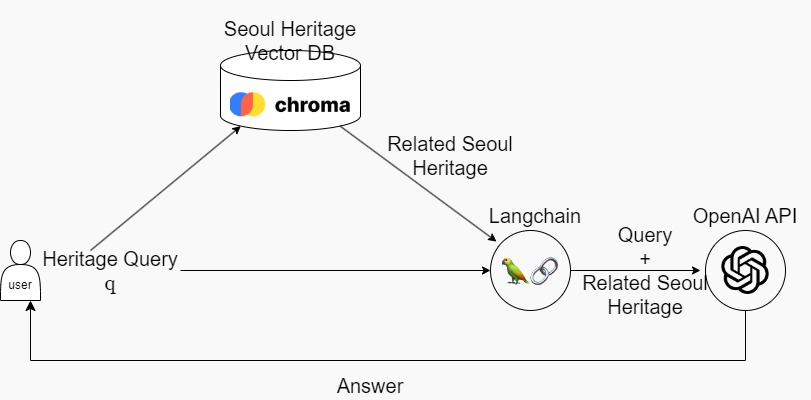} 
\caption{The overall structure of the conversational agent}
\label{fig.1}
\end{center}
\end{figure}

The image illustrates how the agent works. When users ask questions, the system uses both Langchain and the OpenAI API to understand and process these inquiries. It then matches the questions with the relevant information from the given data file and creates a clear answer. This answer is then shown to the user through an interface created with Streamlit. This setup allows the agent to provide immediate and relevant answers, ensuring a smooth and informative conversation with users.

Let's break down the components, starting with the data. The data file has details about heritage sites in Seoul, all written in English. The given data is main source of knowledge, providing names and locations of the heritage sites. This information is then converted into a vector form and is saved in a Chroma database. 

We utilize the GPT-4 model from OpenAI API and set it up using a template from Langchain, giving it a guideline on how it should interact. By combining advanced language model of OpenAI with Langchain, agent can better understand and produce accurate responses. This results in conversations where the agent clearly understands what users are saying and replies in a relevant and meaningful way. 

We've used Streamlit to launch the agent, which gives users a simple and interactive way to communicate with it. Streamlit combines the chat features we built with Langchain and the OpenAI API to create a single platform where users can interact smoothly with the conversational agent. 

We've woven together Streamlit, Langchain, and the OpenAI API to create the foundation of the dialogue system. This combination allows users to easily obtain information about Seoul's heritage sites. Below is a glimpse of what interacting with the agent looks like.

\begin{figure}[!ht]
\begin{center}
\includegraphics[scale=0.4]{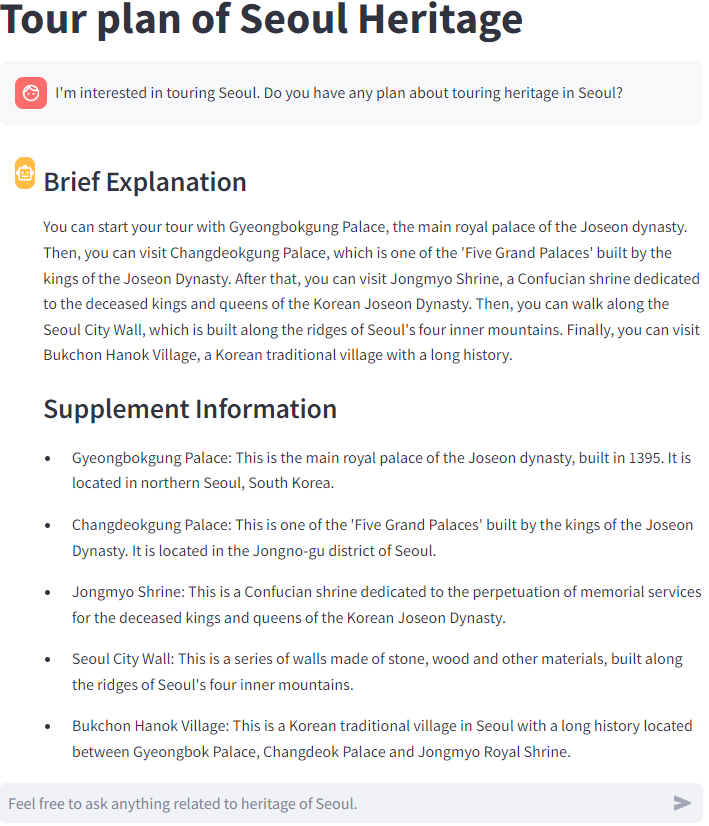} 
\caption{A sample result of the conversation between the user and the agent}
\label{fig.2}
\end{center}
\end{figure}

\section{Discussion}
Our exploration into the chatbot designed to share knowledge about Seoul's historical sites reveals both the strengths of modern technology and the constraints of its data sources. This chatbot creates an engaging experience, making it easier and more approachable for people unfamiliar with Seoul's history to learn about it.

We've made the chatbot more user-friendly by incorporating advanced technologies like Streamlit and Langchain. These tools do more than just make the chatbot look good; they ensure interactions are smooth and easy for everyone. At the heart of this chatbot is a complex natural language understanding system that facilitates conversations that are not only informative but also engaging.

However, there's a catch: the chatbot depends on publicly available data, which has its limits. This reliance means it can only share as much as the data it has access to, which might not cover all of Seoul's historical sites comprehensively. Important places could be missed, or the stories told could be less detailed, affecting the chatbot's effectiveness.

Accuracy is another hurdle. The chatbot's ability to process and respond in natural language can sometimes be hampered by incomplete data, leading to responses that might be too general or incorrect. It's important for users to double-check the information provided by the chatbot, especially if they're looking for detailed and specific insights.

Delving deeper into discussions about particular historical sites presents additional challenges. Our data mainly lists the names and locations of sites. To make conversations flow better with this limited information, we use techniques like crafting prompts that encourage the chatbot to engage with the data in a conversational manner. For example, we might use a prompt like "Make conversation with the dataset," where "dataset" refers to a file containing key information like site names and locations in English, helping to bridge the gap between the data we have and the rich conversations we aim to facilitate.

To generate smoother conversations from this limited data, we employ prompts such as "Make conversation with the dataset" as shown below, where "dataset" denotes a file with keys like "main\_key", "h\_eng\_dong", "h\_eng\_gu", "name\_eng", and "h\_eng\_city". Below is an example about the explanation wrote above.

\begin{figure}[!ht]
\begin{center}
\includegraphics[scale=0.42]{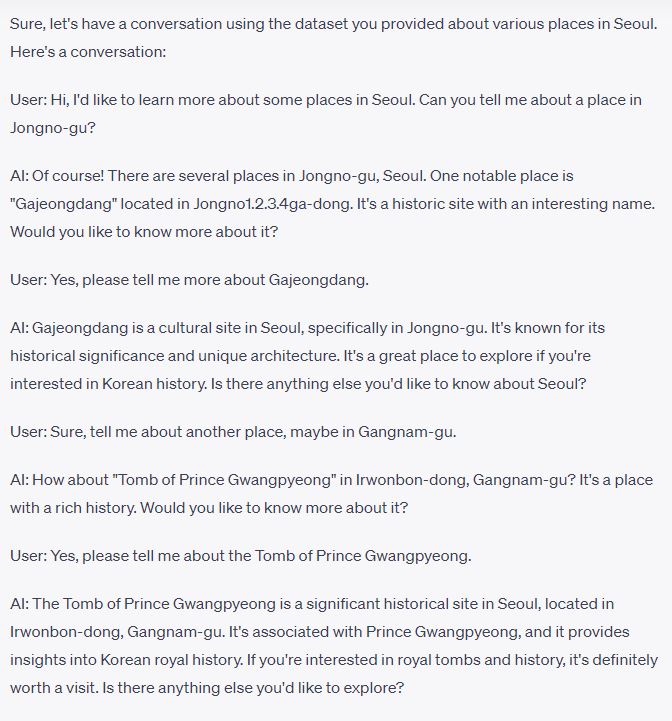} 
\caption{A result of conversation generation with "Make conversation with the dataset" prompt}
\label{fig.3}
\end{center}
\end{figure}

The subsequent interactions, as delineated below, guide users on prompts that elicit more comprehensive discussions. For instance, after an introductory inquiry about Seoul's heritage, the agent can suggest related prompts to enrich the dialogue. So, the user can save time with prompts when continuing the conversation. Still, there remains a risk of truncated or unsatisfying exchanges. To mitigate this, we gave prompts like "Suggest further questions on Seoul's heritage". The results, as showed below, propose relevant questions rooted in the dataset, assisting particularly novice users unsure of the dialogue's progression. 

\begin{figure}[!ht]
\begin{center}
\includegraphics[scale=0.42]{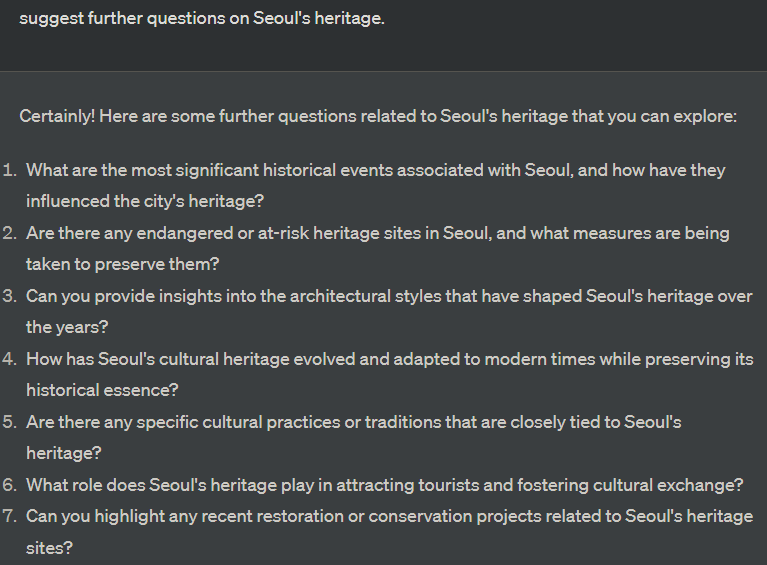} 
\caption{A sample response with "Suggest further questions on Seoul's heritage." prompt}
\label{fig.4}
\end{center}
\end{figure}

\section{Conclusion}

In conclusion, we've developed a prototype of a chatbot designed specifically for those unfamiliar with Seoul's historical landmarks. Utilizing the OpenAI API, integrated with the Langchain framework and Streamlit's interactive development capabilities, we've crafted an agent that not only delivers pertinent information but does so through an intuitive and user-friendly interface.

The synergy of these cutting-edge technologies ensures that the chatbot is both informative and adaptable. The OpenAI API provides a solid foundation for understanding and addressing user inquiries, Langchain helps overcome language barriers, and Streamlit offers a responsive interface that meets the varied needs of its users. While the fusion of these technologies represents a notable accomplishment, we recognize the potential for ongoing enhancements and optimizations.

Reflecting on our journey and the creation of the prototype, we are filled with optimism for its future prospects. Our hope is for this agent to transcend its role as a technological innovation, becoming a symbol of hospitality. We imagine it as a continuously evolving tool, mirroring the vibrant and ever-unfolding story of Seoul's cultural legacy.

\nocite{*}
\section{Bibliographical References}
\label{sec:reference}

\bibliographystyle{lrec-coling2024-natbib}
\bibliography{lrec-coling2024-example}

\section{Language Resource References}
\label{lr:ref}
\bibliographystylelanguageresource{lrec-coling2024-natbib}
\bibliographylanguageresource{languageresource}

\end{document}